\documentclass[journal,twoside]{IEEEtran}
\usepackage{graphicx}
\usepackage{subfig}
\usepackage{amsmath}
\usepackage{bbm}
\usepackage{amssymb}
\usepackage{multirow}
\usepackage{enumerate}
\usepackage[colorlinks,linkcolor=red,anchorcolor=blue,citecolor=green]{hyperref}

\usepackage{cite}
\ifCLASSINFOpdf
\else
\fi
\begin{document}
%\title{Alignment-Free Person Re-Identification by Transformer Decoders}
\title{AAformer: Auto-Aligned Transformer for Person Re-Identification}

\author{Kuan~Zhu,~
	Haiyun~Guo,~\IEEEmembership{Member,~IEEE},~
	Shiliang~Zhang,~\IEEEmembership{Senior Member,~IEEE},~
	Yaowei~Wang,~\IEEEmembership{Member,~IEEE},~\\
	Jing~Liu,~\IEEEmembership{Member,~IEEE},~
    Jinqiao~Wang,~\IEEEmembership{Member,~IEEE},~
    and~Ming~Tang,~\IEEEmembership{Member,~IEEE}
    
\thanks{Manuscript received 5 September 2022; revised 11 May 2023; accepted 24 July 2023. This work was supported in part by the Key-Area Research and Development Program of Guangdong Province under Grant 2021B0101410003; in part by the National Natural Science Foundation of China under Grant 62276260, Grant 62002356, and Grant 61976210; and in part by the Beijing Natural Science Foundation under Grant 4244099. (Corresponding author: Haiyun Guo.)}

\thanks{Kuan Zhu and Ming Tang are with the Foundation Model Research Center, Institute of Automation, Chinese Academy of Sciences, Beijing 100190, China (e-mail: kuan.zhu@nlpr.ia.ac.cn; tangm@nlpr.ia.ac.cn).}

\thanks{Haiyun Guo is with the Foundation Model Research Center, Institute of Automation, Chinese Academy of Sciences, Beijing 100190, China, also with the School of Artificial Intelligence, University of Chinese Academy of Sciences, Beijing 100049, China, and also with the Development Research Institute of Guangzhou Smart City, Guangzhou 510805, China (e-mail: haiyun.guo@nlpr.ia.ac.cn).}

\thanks{Shiliang Zhang is with the National Key Laboratory for Multimedia Information Processing, School of Computer Science, Peking University, Beijing 100871, China (e-mail: slzhang.jdl@pku.edu.cn).}

\thanks{Yaowei Wang is with the Peng Cheng Laboratory, Shenzhen 518066, China (e-mail: wangyw@pcl.ac.cn).}

\thanks{Jing Liu is with the Foundation Model Research Center, Institute of Automation, Chinese Academy of Sciences, Beijing 100190, China, and also with the School of Artificial Intelligence, University of Chinese Academy of Sciences, Beijing 100049, China (e-mail: jliu@nlpr.ia.ac.cn).}

\thanks{Jinqiao Wang is with the Foundation Model Research Center, Institute of Automation, Chinese Academy of Sciences, Beijing 100190, China, also with the School of Artificial Intelligence, University of Chinese Academy of Sciences, Beijing 100049, China, also with the Wuhan AI Research, Wuhan 430073, China, and also with the Peng Cheng Laboratory, Shenzhen 518066, China (e-mail: jqwang@nlpr.ia.ac.cn).}

\thanks{Digital Object Identifier 10.1109/TNNLS.2023.3301856}
}

% Personal use of this material is permitted. However, permission to use this material for any ohter purposes must be obtained from the IEEE by sending a request to pubs-permissions@ieee.org

\markboth{IEEE TRANSACTIONS ON NEURAL NETWORKS AND LEARNING SYSTEMS}%,~Vol.~14, No.~8, August~2018}%
{Zhu \MakeLowercase{\textit{et al.}}: AAformer: Auto-Aligned Transformer for Person Re-Identification}

%\IEEEpubid{Copyright \copyright~2018 IEEE. Personal use is permitted, but republication/redistribution requires IEEE permission.\\
%See http://www.ieee.org/publications\_standards/publications/rights/index.html for more information.}


\maketitle

%However, their straightforward and equal partitions on images make the outcome stripes fixed in position and height, %positions-related but not semantic-related, 
%which cannot align the person images well and introduce lots of background. 
%pixel-level part-aligned features only with person identity labels for person re-ID and make two contributions. 
% by employing the row classifiers to investigate the response of every pixel. 
%The most responsive pixels to human parts are assigned with foreground labels and the most responsive pixels to background part are assigned with background labels. The remaining pixels are considered as neutral pixels and are ignored in training. 
\begin{abstract}
 
In person re-identification, extracting part-level features from person images has been verified to be crucial to offer fine-grained  information. Most of existing CNN-based methods only locate the human parts coarsely, or rely on pre-trained human parsing models and fail in locating the identifiable non-human parts (e.g., knapsack). In this paper, we introduce an alignment scheme in Transformer architecture for the first time and propose the Auto-Aligned Transformer (AAformer) to automatically locate both the human parts and non-human ones at patch-level. We introduce the ``Part tokens ({[PART]}s)'', which are learnable vectors, to extract part features in Transformer. A {[PART]} only interacts with a local subset of patches in self-attention and learns to be the part representation. To adaptively group the image patches into different subsets, we design the Auto-Alignment. Auto-Alignment employs a fast variant of Optimal Transport algorithm to online cluster the patch embeddings into several groups with the {[PART]}s as their prototypes. AAformer integrates the part alignment into the self-attention and the output {[PART]}s can be directly used as part features for retrieval. Extensive experiments validate the effectiveness of {[PART]}s and the superiority of AAformer over various state-of-the-art methods.
 
\end{abstract}


%representations, the average pooling results of rows on feature maps,

% \fi
\IEEEpeerreviewmaketitle

\section{Introduction}

%The self-attention based architecture, Transformer \cite{Transformer}, which is regarded as the new standard in natural language processing (NLP), has recently shown its scalability and effectiveness in many computer vision tasks. Deosovitskiy \etal \cite{ViT} apply a typical Transformer to image recognition with the fewest modifications. The proposed Vision Transformer (ViT) obtains better performance and generalization than traditional convolutional neural networks (CNNs).

%Deosovitskiy \etal \cite{ViT} attempt to apply a typical transformer to image recognition with fewest adjustments and propose the Vision Transformer (ViT) which provides better performance than a traditional convolution neural network.

\begin{figure}[t]
\begin{center}
\includegraphics[width=1.0\linewidth]{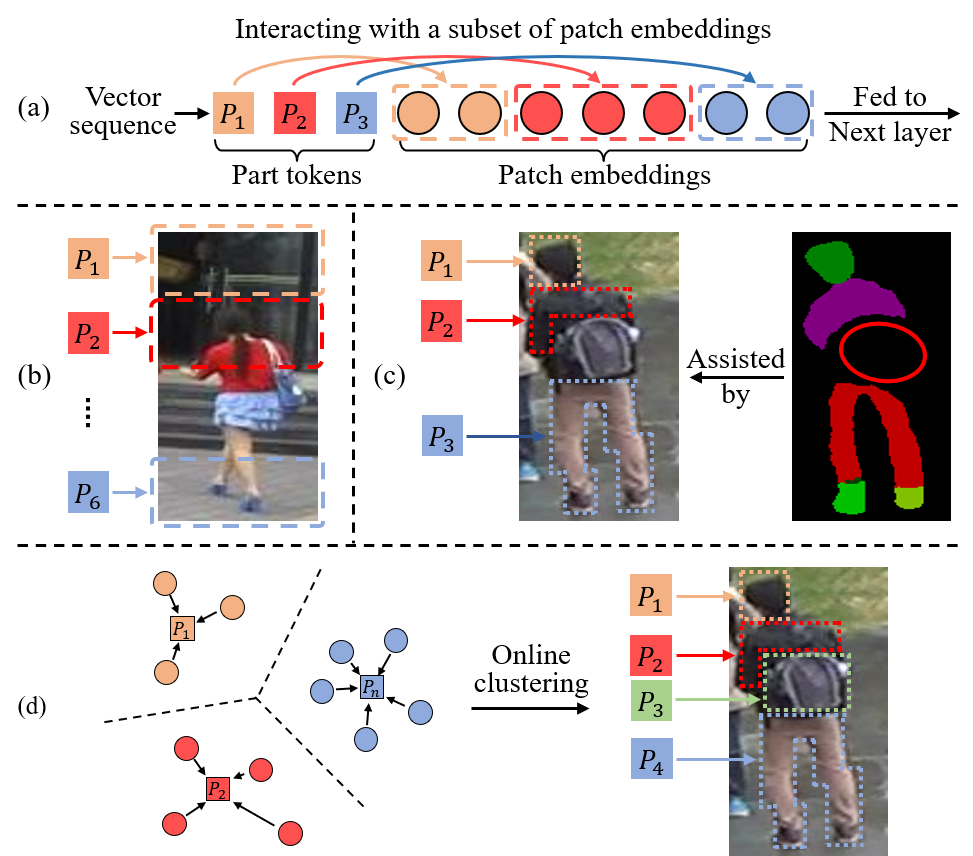}
\end{center}
   \caption{(a) The illustration of the added {[PART]}s. {A {[PART]} only interacts with a subset of the patch embeddings and thus can learn to represent the subset.} (b) {[PART]}s with PCB's partitioning~\cite{PCB}. (c) {[PART]}s with SPReID's partitioning~\cite{SPReID}. (d) A toy example of the proposed Auto-Alignment mechanism. {The image patches of the same part, which can be a human part or non-human one, are adaptively grouped to the identical {[PART]}.}}
\label{fig:introduction}
\end{figure}

%(b) Part tokens applied to PCB~\cite{PCB}. (c) Part tokens applied to MGN~\cite{MGN}.

%(b) PCB~\cite{PCB} based on part tokens. (c) MGN~\cite{MGN} based on part tokens.

%However, in person re-identification (re-ID) task, which aims to associate the specific person across cameras, convolutional architectures remain dominant \cite{AlignedReID, DSA-reID, PCB, MGN, SPReID, ISP}. Encouraged by the Transformer scaling successes in various vision tasks, we wonder whether the Transformer can handle more fine-grained and difficult tasks like person re-ID. Therefore, in this paper, we propose to apply the pure Transformer architecture to person re-ID to obtain higher accuracy than CNN-based methods. 

Person re-identification (re-ID) aims to associate the specific person across cameras. Lots of existing convolution neural network-based methods extract part-level features to obtain fine-grained information to alleviate the body part misalignment problem, which can be caused by inaccurate person detection, human pose variations, and the changing of camera viewpoints. {The stripe-based methods \cite{PCB,MGN} design stripe-based image partitions to locate human parts.} Some methods adopt pre-trained human parsing models \cite{SPReID} to locate human parts of fine granularity. These methods either cannot well align the human parts or fail in locating the identifiable non-human parts like knapsack \cite{ISP}.

%which is regarded as the new standard in natural language processing (NLP),

Recently, the self-attention-based architecture, Transformer \cite{Transformer}, has shown its scalability and effectiveness in many computer vision tasks. ViT \cite{ViT} first employs Transformer architecture to conduct image recognition. They divide the input image into fixed-size patches, linearly embed each of them, and fed them to a Transformer encoder to obtain the image representation. TransReID \cite{transreid} proposes to apply the ViT to object re-identification. They design patch shuffle operations and introduce side information like camera/view IDs to learn robust features, {validating the superiority of Transformer in re-ID tasks}. As they do not take the misalignment issue into consideration, the part alignment scheme in the Transformer architecture is still unexplored. %how to explicitly extract discriminative part features for person images is still blank in Transformer architecture.

In this paper, we propose a specific alignment scheme for Transformer and also alleviate the problems of existing methods, i.e., aligning human parts coarsely and failing in locating non-human parts. The proposed Auto-Aligned Transformer (AAformer) integrates the part alignment into the self-attention and adaptively locate both the human parts and non-human ones at patch-level in an online manner.

%The major challenge in such application to the person re-ID lies in how to extract robust representations of person images with the inevitable problem of body part misalignment caused by inaccurate person detection, human pose variations, or the changing of camera viewpoints. Experimentally, we found that it generated poor accuracy by naively applying the ViT \cite{ViT} to the person re-ID due to the lacking of explicit alignment mechanism. Unfortunately, we also found that the alignment paradigm of CNNs \cite{PCB, MGN, SPReID}, which adopts part masks with part pooling to obtain part features, is ineffective to Transformer. The reason is that the pooling is not suitable for Transformer. Unlike CNN, whose convolutional kernels will highlight specific features, Transformer does not enhance or suppress any information in the self-attention. Employing pooling to Transformer would confuse all kinds of useful or disturbing information, resulting in the poor capability to represent an image. Therefore, a specific alignment framework for Transformer is desirable.

%In view of this, we abandon the pooling operation and introduce the ``part tokens'' to Transformer to learn part representations.
% which is shown in Figure~\ref{fig:introduction}.
%We add the extra learnable part tokens to the vector sequence of image patches and feed the outcome to the Transformer.

First, we propose the ``Part tokens ({[PART]}s)'', which are learnable vectors, for Transformer to learn the representations of local parts. Several [PART]s are concatenated to the sequence of patch embeddings and subsequently fed to the Transformer encoder. In self-attention, a {[PART]} only attends to a local subset of patch embeddings and learns to represent the subset, which is shown in the first row of Figure~\ref{fig:introduction}. The existing CNN-based methods can be transferred to Transformer by {[PART]}s, e.g., PCB \cite{PCB} and SPReID \cite{SPReID}, which are illustrated in the second row of Figure~\ref{fig:introduction}, showing the problems of aligning human parts coarsely and failing in locating  non-human parts.

By means of the {[PART]}, we propose the Auto-Aligned Transformer (AAformer) to adaptively group the image patches into different subsets. We employ a fast variant of Optimal Transport algorithm \cite{Sinkal} to online cluster the patch embeddings into several groups with the {[PART]}s as their prototypes (e.g., leg prototype, knapsack prototype). The patch embeddings of the same semantic part are gathered to the identical {[PART]} and the {[PART]} only interacts with these patch embeddings. Given an input image, the output {[PART]}s of AAformer learn to be different part prototypes of patch subsets of the input image, and can be directly used as part features for re-ID. A toy example of this Auto-Alignment is shown in the last row of Figure~\ref{fig:introduction}. In each layer of AAformer, we sequentially conduct the Auto-Alignment to align the body parts and the Self-Attention to learn the part representations in the feedforward propagation. Therefore, AAformer is  an  online  method that simultaneously learns both part alignment and part representations, locating both human parts and non-human ones more accurately.

The contributions of this work are summarized as follows:

1) We introduce the ``Part tokens ({[PART]}s)'' for Transformer to learn part features and integrate the part alignment into the self-attention. A {[PART]} only interacts with a subset of patch embeddings, {and thus can learn to be the local representation.} %The part alignment is conducted throughout the model with self-attention.

%of the same semantics to the specific part token to that semantics

2) {We further propose the AAformer to adaptively locate both the human parts and non-human ones online.} Instead of using a pre-defined fixed patch assignment, AAformer simultaneously learns both part alignment and part representations. 

3) Extensive experiments validate the effectiveness of {[PART]} and the superiority of AAformer over various state-of-the-art methods on the widely used benchmarks, i.e., Market-1501 \cite{Market1501}, CUHK03 \cite{CUHK03-1}, DukeMTMC-reID \cite{DukeMTMC-reID} and MSMT17 \cite{MSMT17}.

The rest of this paper is arranged as follows. The second
part reviews some works related to this paper. The third part
introduces the main architecture of the proposed method and expounds the proposed method in detail. The fourth part shows the good performance of the proposed method, compares it with the state-of-the-art methods, and
analyzes the reasons why the proposed method is effective.
The fifth part summarizes the full paper.

\section{Related Work}

\subsection{Aligned Person Re-identification}

Lots of existing CNN-based person re-ID approaches focus on global representation learning of identity-discriminative information, e.g., IDE network learning \cite{deeplearning1, CUHK03-1,  deeplearning2, deeplearning3, deeplearning4, deeplearning5}, metric learning \cite{metric1, metric2, metric3, metric4, metric5, metric6}. However, the lacking of explicit alignment mechanism largely limits their performance. To remedy this, many efforts have been made to develop the aligned person re-ID, which aims to extract part-level features. These methods could be roughly summarized to the following streams:

1) \textbf{Stripe-based approaches.} Some researchers develop stripe-based methods to extract part features. They directly partition the person images into several fixed horizontal stripes. PCB \cite{PCB} first equally partitions the person images and then adopts part
classifiers to refine each part in an attention-like manner. MGN \cite{MGN} enhances the robustness by dividing images into stripes of different granularities and designs overlap between stripes. However, these stripes are too coarse and with fixed heights and positions but do not correspond to specific semantic parts, and therefore fail in aligning different human parts well. Besides, there still remains much background noise in their stripes. Compared with these methods, our AAformer can adaptively locate different local parts at patch-level, which is more accurate.

2) \textbf{Bounding box-based Approaches.} Some works propose to locate the local parts by incorporating a bounding box selection sub-network \cite{MSCAN, DPL, PAR}. MSCAN \cite{MSCAN} employs the STN \cite{STN} to locate the latent discriminative parts and subsequently extracts the part-level features. DPL \cite{DPL} proposes to generate bounding boxes  from saliency maps and then extract part-level features from these parts. However, {the shape of located areas of these methods is limited to the rectangle,} thus the located human parts are still coarse. Besides, {there is usually much overlap between the bounding boxes predicted by these methods}. In contrast, {AAformer can adaptively locate different discriminative parts with arbitrary shapes and ensure that there is no overlap between each part.}

3) \textbf{Extra semantics-based methods.} Some other works employ extra semantics to locate human parts. SPReID \cite{SPReID} proposes to employ a pre-trained human semantic parsing model to provide the mask of body parts for alignment. PS \cite{PS} adpots the human segmentation model to conduct the part awareness learning process. CDPM \cite{CDPM} also proposes to use the human parsing results to align the human body part in the vertical direction.  DSA-reID \cite{DSA-reID} adopts dense extra semantic information of 24 regions for a person. However, the identifiable personal belongings like knapsack and reticule, which are crucial for person re-ID, cannot be recognized by the pre-trained models and {are ignored as background}. Besides, the accuracy of extra semantics heavily counts on the pretrained human parsing models or the pose estimators. And these approaches cannot re-correct the errors of semantic estimation in their training processes. {In contrast, AAformer can adaptively locate all the identifiable parts including both the human parts and nonhuman ones and the adaptive patch assignment is conducted online in every layer of AAformer, so even if some patches are assigned incorrectly in one AAformer layer, other layers can also make the right patch assignment, which increases the robustness.}

4) \textbf{Attention-based methods.} Attention mechanism constructs alignment by suppressing background noise and enhancing the discriminative regions \cite{MHN,HA-CNN,DuATM,Mancs,CAMA,CASN}. HA-CNN \cite{HA-CNN} formulates the harmonious attention CNN model by joint learning of soft pixel attention and hard regional attention. MHN \cite{MHN} utilizes the complex and high-order statistics information in {the attention mechanism}, so as to capture the subtle but discriminative foreground areas in person images. However, the part partition of these methods is also coarse and these methods cannot explicitly locate the semantic parts. Besides, the semantic consistency of {the focus area} among images is not guaranteed. By contrast, our AAformer can locate the identifiable local parts at {the patch-level} and  guarantee the semantic consistency among images by the proposed {[PART]}s. 
%In this paper, we propose an online method, Auto-Aligned Transformer (AAformer), which can adaptively locate both the human parts and non-human ones at patch-level without the help of extra semantics. 
Apart from AAformer, ISP \cite{ISP} also proposes to locate both human parts and non-human ones automatically by iterative clustering, but their off-line part partition is a little bit time-costing and prevents their method from end-to-end training. % Besides, they use part pooling to generate part features, and this way is not suitable for Transformer as pooling is usually not applied in Transformer \cite{ViT, Bert}.  % In general, these CNN-based alignment methods can be summarized as part masks (either designed artificially or provided by pre-trained models) with part pooling. For example, the stripe-based methods \cite{PCB, MGN} design the stripe-based part masks on global features and employ part pooling on each stripe. The extra semantics-based methods \cite{SPReID, DSA-reID} adopt the pixel-level part masks and conduct part pooling on each part. However, we experimentally verify that this paradigm, which is effective for CNNs, is ineffective for Transformer. This phenomenon motivates us to design a specific alignment framework for Transformer.

%However, these stripes are with fixed heights and positions but not corresponding to specific semantic parts, thus fail in aligning different human parts well. Besides, there still remains much background noise in their stripes. Compared with these methods, SCS+ can learn the semantics-consistent stripes and further remove the background noise.

%\textbf{Extra semantics based approaches.} Many works employ extra semantics in terms of part/pose to locate different body parts \cite{P2-Net, PoseTransfer, PGFA, PSE, MGCAM, PDC, AANet, GLAD, AACN, PoseInvariant}. Kalayeh et al. \cite{SPReID} propose to employ a pre-trained human semantic parsing model to achieve the pixel-level part alignment. Zhang et al. \cite{DSA-reID} further adopt DensePose \cite{DensePose} to get densely semantics of 24 regions for a person. However, the requirement of extra semantics extremely limits their utility and robustness. First, the accuracy of extra semantics heavily counts on the pre-trained human parsing models or the pose estimators. And these approaches cannot recorrect the errors of semantic estimation in their training processes. Second, beyond the pre-defined human parts, there still exist many identifiable non-human parts like backpacks and reticule, which could be critical for person re-ID, while recognized as background by the pre-trained human parsing models. Compared with them, SCS+ can locate both the identifiable human parts and non-human parts at pixel-level without the need of extra semantics.

\begin{figure*}[t]
\begin{center}
\includegraphics[width=1.\linewidth]{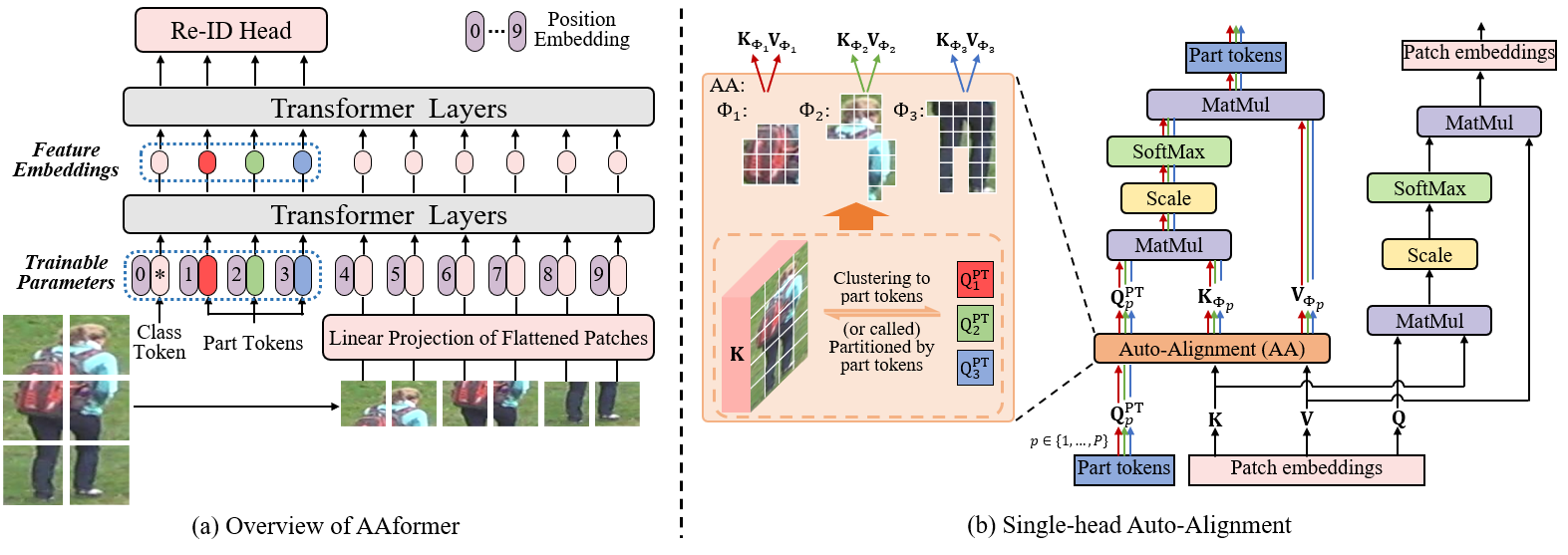}
\end{center}
   \caption{(a) The overview of AAformer. We divide the input image into fixed-size patches, linearly embed each of them and add the position embeddings. We add the extra learnable vectors of ``Class token (CLS)'' and ``Part tokens ({[PART]}s)'' to learn the global and part representations of person images. The {[PART]}s fed to the first Transformer layer are parameters of AAformer which learn the part prototypes of the datasets. The {[PART]}s output by Transformer layers are learned feature embeddings to represent human parts for input images. (b) Single-head Auto-Alignment. The self-attention for {[PART]}s is replaced by Auto-Alignment. $\mathbf{Q}^{\rm PT}$: \emph{query} vectors of {[PART]}s. $\Phi_p$: the patches assigned to the $p$th {[PART]}. $\mathbf{Q, K, V}$: \emph{query}, \emph{key}, \emph{value} vectors of patch embeddings. %The processing for CLS is the same as patch embeddings and CLS is omitted for simplicity.
   }
\label{fig:overview}
\end{figure*}

\subsection{Visual Transformer}
%like object detection \cite{DETR}, generative adversarial network \cite{TransGAN}, action recognition \cite{TREAR}.

%Transformer is proposed by \cite{Transformer} for machine translation and has since become the state-of-the-art architecture in many NLP tasks. 
Recently, the Transformer \cite{Transformer} is showing its superiority over conventional methods in many vision tasks. ViT \cite{ViT} proposes the Vision Transformer to apply a pure Transformer to image recognition. They first divide the input image into image patches and then map them to a sequence of vectors with a linear projection. An extra learnable ``class token (CLS)'' is added to the sequence and the vector sequence is fed to a typical Transformer encoder. TransReID \cite{transreid} proposes to apply ViT to object re-identification. They design patch shuffle operations and introduce side information like camera/view IDs to learn robust features, validating the superiority of the Transformer on re-ID task. {DCAL \cite{DCAL} proposes to implicitly extract the local features using a Transformer decoder.} PAT \cite{PAT} and HAT \cite{HAT} both integrate Transformer architecture into CNNs. PAT proposes to use Transformer to generate the attention masks for CNN and HAT uses Transformer to aggregate hierarchical features from CNN backbones. However, {as they do not design an explicit alignment scheme for Transformer,} how to extract discriminative part features from person images is still unexplored in Transformer architecture. In this paper, {we aim to propose a specific alignment scheme} for Transformer and also 
alleviate the problems in existing CNN-based alignment methods.

{Besides, some methods \cite{GroupViT, VPT} also design to add learnable tokens to the Transformer network. GroupViT \cite{GroupViT} uses the Gumbel Softmax function to group the patches to the learnable group tokens, which assign the image patches to their most similar group tokens with a high probability. Compared with GroupViT, AAformer uses Optimal Transport to assign image patches to [PART]s, which can avoid the trivial solution where all the patches are assigned to the same [PART]. VPT \cite{VPT} fixes the parameters of a pre-trained model and only trains the added learnable token to reduce the number of parameters and calculation cost. The learnable token in VPT still interacts with all the image patches. While in AAformer, a [PART] only interacts with a subset of patch embeddings, thus can learn to be the local representation.}

\section{Methodology}

%We begin by briefly revisiting the well-established Vision Transformer architecture \cite{ViT}. 
In this section, {we expound the proposed method in detail}. We begin by briefly revisiting the general Transformer architecture in Sect. \ref{arch}. 
Then we present the proposed {[PART]} and the Auto-Aligned Transformer (AAformer) in Sect. \ref{AAformer} step by step. %Lastly, we introduce the objective function used in our experiments.

\subsection{The Main Architecture}\label{arch}
%The overview of AAformer is shown in Figure~\ref{fig:overview}, of which the main structure is Vision Transformer (ViT) \cite{ViT}. 
We follow Vision Transformer (ViT) \cite{ViT} to construct the main architecture.  
Given an input image $\mathbf{x}\!\in\!\mathbb{R}^{H\times W\times C}$ with resolution $(H,W)$ and channel $C$, we reshape it into a sequence of flattened 2D patches $\mathbf{x}\in\mathbb{R}^{N\times(I^2\cdot C)}$ to fit the Transformer architecture, where $(I,I)$ is the resolution of each image patch and $N=(H\cdot W)/I^2$ is the length of image patch sequence. We map the patches to vectors of $D$ dimensions with a trainable linear projection and refer to the output of this projection as the patch embeddings. A standard embedding of ``class token (CLS)" is added to extract the global representation. Lastly, the outcome vector sequence $\mathbf{Z}\in\mathbb{R}^{L\times D}$ is fed to the Transformer encoder, where $L\!=\!1\!+\!N$. We also add the standard learnable 1D position embeddings to the vector sequence in element-wise to retain positional information.

The standard Transformer layer \cite{Transformer} consists of Multi-head Self-Attention (MSA) and Multi-Layer Perception (MLP) blocks. 
The self-attention mechanism is based on a trainable associative memory with $key$ and $value$ vector pairs. For every $query$ vector in a sequence of query vectors ($\mathbf{Q}\in\mathbb{R}^{L\times D}$), we calculate its inner products with a set of $key$ vectors ($\mathbf{K}\in\mathbb{R}^{L\times D}$). These inner products are then scaled and normalized with a softmax function to obtain $L$ weights. The output of the self-attention for this $query$ is the weighted sum of a set of $L$ $value$ vectors ($\mathbf{V}\in\mathbb{R}^{L\times D}$). For all the $queries$ in the sequence, the output matrix of self-attention can be obtained by:
\begin{equation}\label{self_attention}
Attention(\mathbf{Q}, \mathbf{K}, \mathbf{V})=Softmax(\frac{\mathbf{Q}\mathbf{K}^T}{\sqrt{D}})\mathbf{V},
\end{equation}
where the $Softmax$ function is applied over each row of the input matrix and the $\sqrt{D}$ term provides appropriate normalization. $Query$, $key$ and $value$ matrices are all computed from the vector sequence $\mathbf{Z}$ using different linear transformations: $\mathbf{Q}\!=\!\mathbf{Z}\mathbf{W}_Q,~\mathbf{K}\!=\!\mathbf{Z}\mathbf{W}_K, ~\mathbf{V}\!=\!\mathbf{Z}\mathbf{W}_V$. Finally, the Multi-head Self-Attention layer (MSA) is defined by considering $h$ attention ``heads'', i.e., $h$ self-attention functions are applied to the input in parallel. Each head provides a sequence of size $L\times d$, where $d=D/h$ typically. The outputs of the $h$ self-attention are rearranged into a $L\times D$ sequence to feed the next Transformer layer.

The MSA is followed by a MLP block to build a full Transformer layer. This MLP contains two linear layers separated by a GELU non-linearity \cite{GELU}. The first linear layer expands the dimension from $D$ to $4D$ and the second reduces the dimension back to $D$. Both MSA and MLP are operating as residual connections and with a layer normalization (LN) \cite{LN} before them. \textbf{We change the MSA layer to Multi-head Auto-Alignment (MAA) layer to build our AAformer.} The overview of AAformer is shown in the left part of Figure~\ref{fig:overview}.

\subsection{Auto-Aligned Transformer}\label{AAformer}

%We add $P$ learnable embeddings of ``part tokens'' to the sequence of patch embeddings to extract the part representations for the input image.

\textbf{{[PART]}s for Transformer.} We propose the {[PART]}s for Transformer to extract the part features and integrate the part alignment into self-attention. We concatenate $P$ {[PART]}s, which are learnable vectors, to the sequence $\mathbf{Z}$, thus now the length $L$ of $\mathbf{Z}$ is $1\!+\!P\!+\!N$. A {[PART]} only interacts with a subset of patch embeddings, rather than all of them, {and thus can learn to be the partial representation of the subset.} Specifically, we denote $\mathbf{Q}^{\rm PT}\!=\![\mathbf{Q}^{\rm PT}_1,...,\mathbf{Q}^{\rm PT}_P]$ as the $query$ vectors of {[PART]}s and denote $\Phi_p$ as the subset of patch embeddings assigned to the $p$th {[PART]}. For the $query$ vector of the $p$th {[PART]}, denoted as $\mathbf{Q}^{\rm PT}_p$, we only calculate its inner products with the $key$ vectors belonging to $\Phi_p$ and then scale and normalize these inner products with softmax function to obtain $len(\Phi_p)$ weights. The output of the part alignment for $\mathbf{Q}^{\rm PT}_p$ is the weighted sum of \emph{value} vectors belonging to $\Phi_p$, which is formulated as:
\begin{equation} \label{Part}
Alignment(\mathbf{Q}^{\rm PT}_p, \mathbf{K}, \mathbf{V})=Softmax(\frac{\mathbf{Q}^{\rm PT}_{p}\mathbf{K}_{\Phi_{p}}^T}{\sqrt{D}})\mathbf{V}_{\Phi_p},
\end{equation}
where $\mathbf{K}_{\Phi_{p}}$ and $\mathbf{V}_{\Phi_p}$ are the $key$ and $value$ vectors belonging to $\Phi_{p}$, respectively.
After the part alignment, the output vector of {[PART]} $p$ becomes the part representation of subset $\Phi_{p}$. The CLS token and patch embeddings are processed by the original self-attention (Eq.~\ref{self_attention}). 

Now, to extract discriminative part features, the core problem is how to accurately and adaptively assign image patches to ${\Phi_p}, p\!\in\{\!1,...,P\}$.

\textbf{Multi-head Auto-Alignment.} The Auto-Alignment aims at automatically grouping the image patches into different $\Phi_{p}~ (p\!\in\{\!1,...,P\})$, in which both human parts and non-human ones are included. The $\Phi_{p}$ for different {[PART]}s should be mutually exclusive. Our idea is adaptively clustering the patch embeddings into $P$ groups with the {[PART]}s as their prototypes. This makes the patch assignment problem the same as the Optimal Transport problem \cite{Sinkal}. In detail, the $query$ vectors of {[PART]}s, $\mathbf{Q}^{\rm PT}\in\mathbb{R}^{P \times D}$, are regarded as the part prototypes, and we are interested in mapping the $key$ vectors of image patches $\mathbf{K}\in\mathbb{R}^{N\times D}$ to the $\mathbf{Q}^{\rm PT}$ to obtain the patch assignment (CLS token does not participate in the clustering and is omitted for simplicity).
We denote this mapping by $\mathbf{Y}\in\mathbb{R}^{P\times N}$, and the value in position $(p,n)$ denotes the probability of the $n$th image patch belonging to the $p$th {[PART]}. To maximize the similarity between $\mathbf{K}$ and $\mathbf{Q}^{\rm PT}$, $\mathbf{Y}$ is optimized by:
\begin{equation} \label{solve}
    \max_{\mathbf{Y} \in \mathcal{Y}} \operatorname{Tr}\left(\mathbf{Y}^{\top} \mathbf{Q}^{\rm PT} \mathbf{K}^{\top}\right)+\varepsilon H(\mathbf{Y}),
\end{equation}
where $\operatorname{Tr}\left(\mathbf{Y}^{\top} \mathbf{Q}^{\rm PT} \mathbf{K}^{\top}\right)$ measures the similarity between $\mathbf{K}$ and $\mathbf{Q}^{\rm PT}$, {$\operatorname{Tr}$ means the trace of a matrix,} $H(\mathbf{Y})\!=\!-\!\sum_{i j}\mathbf{Y}_{i j} \log\mathbf{Y}_{i j}$ is the entropy function, and $\varepsilon$ is the parameter that controls the smoothness of the mapping and is set to be 0.05 in our experiments. 

To further prevent the trivial solution where all the patches are assigned to the same {[PART]}, we enforce an equal partition by constraining the matrix $\mathbf{Y}$ to belong to the transportation polytope \cite{SeLa,SwAV}. We adapt this constraint to patch assignment by restricting the transportation polytope to the image patches of an image:
\begin{equation}
    \mathcal{Y}=\left\{\mathbf{Y} \in \mathbb{R}_{+}^{P \times N} \mid \mathbf{Y} \mathbbm{1}_{N}=\frac{1}{P} \mathbbm{1}_{P}, \mathbf{Y}^{\top} \mathbbm{1}_{P}=\frac{1}{N} \mathbbm{1}_{N}\right\},
\end{equation}
where $\mathbbm{1}_{P}$ denotes the vector of ones in dimension $P$. These constraints enforce that on average each {[PART]} is selected at least ${N}/{P}$ times in an image. The soft assignment $\mathbf{Y}^*$ are the solution to Prob.~\ref{solve} over the set $\mathbf{\mathcal{Y}}$ and takes the form of a normalized exponential matrix \cite{Sinkal}:
\begin{equation}
    \mathbf{Y}^{*}=\operatorname{Diag}(\mathbf{u}) \exp \left(\frac{\mathbf{Q}^{\rm PT} \mathbf{K}^{\top}}{\varepsilon}\right) \operatorname{Diag}(\mathbf{v}),
\end{equation}
where $\mathbf{u}$ and $\mathbf{v}$ are renormalization vectors in $\mathbb{R}^{P}$ and $\mathbb{R}^{N}$ respectively, {$\operatorname{Diag}(*)$ represents a matrix with $*$ as its diagonal elements}. These renormalization vectors are computed using a small number of matrix multiplications by a fast variant of Sinkhorn-Knopp algorithm \cite{Sinkal}. This algorithm can be efficiently run on GPU and we observe that {only 3 iterations are sufficient to obtain good performance} for patch assignment. In our experiments, grouping 576 patch embeddings of an image to 5 {[PART]}s only takes 0.46ms.
Once a continuous solution $\mathbf{Y}^*$ is found, the patch assignment can be obtained by using a rounding procedure. {The patch embeddings mapped to $\mathbf{Q}^{\rm PT}_p$ form the patch subset $\Phi_p$.} We illustrate the Auto-Alignment process (single-head) in the right part of Figure~\ref{fig:overview}. {The Auto-Alignment is conducted in parallel in different attention heads}, which enhances the robustness of patch assignment. We name this Multi-head Auto-Alignment (MAA) and the outputs of different attention heads are concatenated together. 
The output {[PART]}s of AAformer are used to perform person re-ID.

{In each layer of AAformer, we conduct Auto-Alignment to align body parts and the Self-Attention to learn part representations in the feedforward propagation. Therefore, AAformer is an online and end-to-end method that simultaneously learns both part alignment and part representations.}

\begin{figure}[t]
\begin{center}
\includegraphics[width=1.\linewidth]{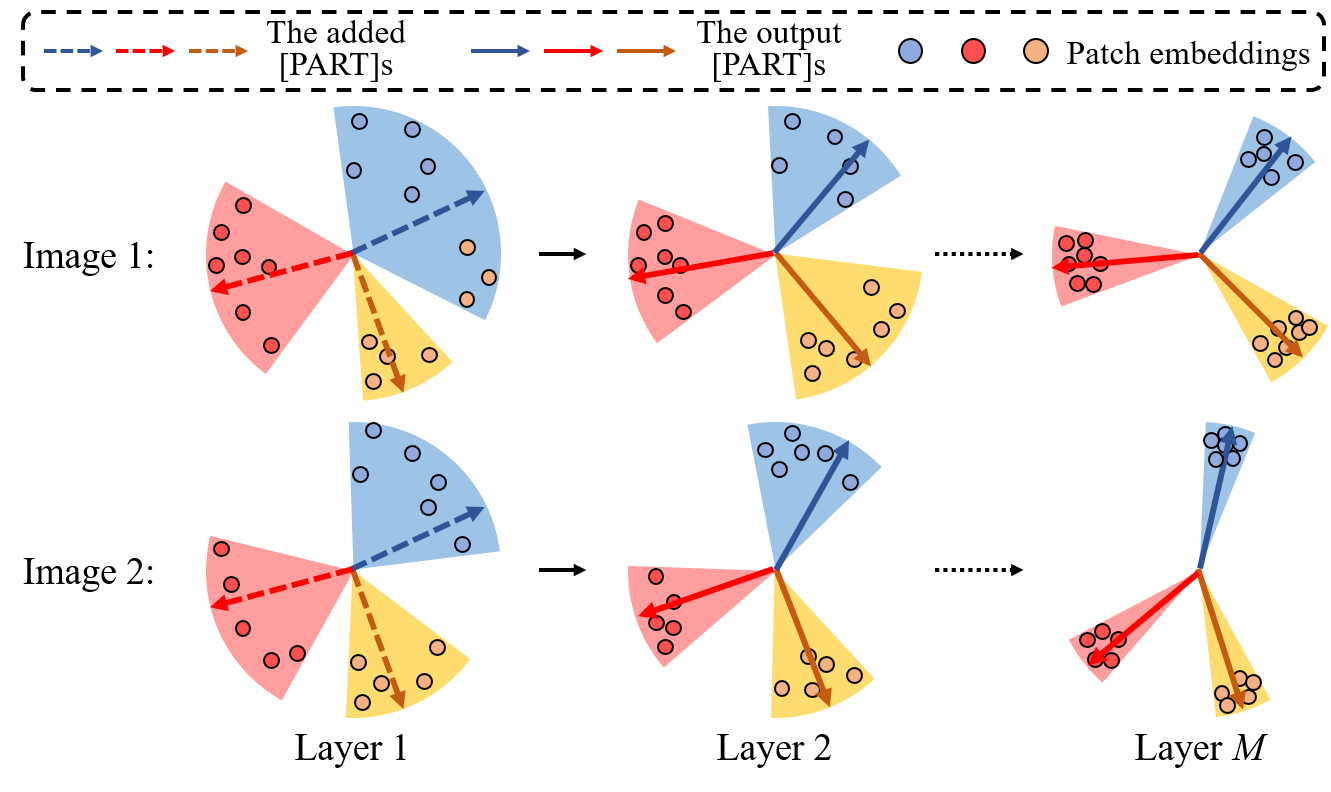}
\end{center}
   \caption{{Illustration of [PART] in different layers of AAformer. The [PART]s fed to the first Transformer layer are the learnable parameters we add to the network. They will learn to be the part prototypes of the dataset in the training and are dataset-adaptive. The output [PART]s of the Transformer layers are the part representations of the input images and they are instance-adaptive.}
   }
\label{fig:OT}
\end{figure}

{{\bf Discussion on [PART]s in AAformer.} (1) The [PART]s fed to the first Transformer layer are learnable parameters we add to the network, which are shown as dashed lines in Figure~\ref{fig:OT}. After continually interacting with the patch embeddings of the same semantic part of different person identities during the training, the added [PART]s learn to be the part prototypes of the training dataset and are dataset-adaptive. This also guarantees the semantic consistency throughout the dataset. (2) After the [PART]s go through the Transformer layers and interact with the patch embeddings of an input image, the output [PART]s of Transformer layers are the part representation of patch subsets of the input image, which are shown as solid lines in Figure~\ref{fig:OT}. That is, the output [PART]s of each Transformer layer are instance-adaptive and can be employed directly as part-level features of the input image for the re-ID task.}

\subsection{Objective Function}
In the training phase, the re-ID heads \cite{baseline} are attached to the output CLS and {[PART]}s of AAformer. Specifically, the output CLS $\mathbf{Z}_0$ represents the global feature of the input image, and the output {[PART]}s $\mathbf{Z}_{1:P}$ are the part features. We employ cross-entropy loss and triplet loss with hard sample mining \cite{triplet_loss} to train our model. The cross-entropy loss is calculated as:
\begin{equation}
    L_{cls}=\frac{1}{P+1} \sum_{i=0}^{P}-\log \mathcal{P}({\mathbf{Z}_{i}}),
\end{equation}
where $\mathcal{P}({\mathbf{Z}_{i}})$ is the probability of token $\mathbf{Z}_{i}$ belonging to its ground truth identity. The classifiers of different tokens are not shared. Label smoothing \cite{label_smooth} is adopted to improve the performance.

The {[PART]}s $\mathbf{Z}_{1:P}$ are concatenated together to calculate the part triplet loss. Together with the global triplet loss calculated with CLS, there are two terms in the triplet loss:
\begin{equation}
    L_{tri}=\frac{1}{2}\left([d^g_p-d^g_n+\alpha]_{+}+[d^p_p-d^p_n+\alpha]_{+}\right),
\end{equation}
where $d_p^g$ and $d_n^g$ are feature distances of global representation (CLS) from positive pair and negative pair, respectively. $d_p^p$ and $d_n^p$ are feature distances of concatenated {[PART]}s from positive pair and negative pair, respectively. $\alpha$ is the margin of triplet loss and is set to 0.3 . $[\cdot]_+$ equals to $max(\cdot, 0)$. {The triplet losses are calculated with hard sample mining strategy [47]. Specifically, for each image, the hard sample mining strategy only uses the hardest positive sample (the least similar positive sample) and the hardest negative sample (the most similar negative sample) to form the positive pair and negative pair. In AAformer, the triplet losses for global feature and part feature are calculated separately. That is, the hardest positive and negative pairs found by global and part features can be different. Therefore, there are two groups of positive and negative pairs for each image in AAformer.} Therefore, the overall objective function for our model is:
\begin{equation}
    L=L_{cls}+L_{tri}.
\end{equation}

In the testing phase, CLS and {[PART]}s are concatenated together to represent a person image.

\section{Experiments}

In this section, we first list the implementation details of the proposed method. Then, we compare the proposed method with the state-of-the-art methods on both holistic and occluded person re-ID benchmark. At last, we conduct the ablation studies including the effectiveness of {[PART]}s, the effectiveness of AAformer, the comparison between Optimal Transport and Nearest Neighbor, and the visualization of AAformer. All these results validate the effectivenss of the proposed method.

%\subsection{Datasets and Evaluation metrics}
%{\bf Holistic person re-ID Datasets.} We choose four widely used holistic person re-ID benchmarks for evaluation. %, i.e., Market-1501 \cite{Market1501}, DukeMTMC-reID \cite{DukeMTMC-reID} and CUHK03 (New Protocol) \cite{CUHK03-1, CUHK03-2}.
%The Market-1501 \cite{Market1501} contains 19732 gallery images and 12936 training images from 1501 identities captured by 6 cameras. The DukeMTMC-reID \cite{DukeMTMC-reID} contains 1404 identities, 16522 training images, 2228 queries, and 17661 gallery images. The CUHK03 \cite{CUHK03-1, CUHK03-2} contrains 14097 person iamges of 1467 identities and offers both hand-labeled and DPM-detected \cite{DPM} bounding boxes. We adopt the new training/testing protocol proposed in \cite{CUHK03-2}. The MSMT17 \cite{MSMT17}

%{\bf Occluded person re-ID dataset.} We also evaluate the performance of AAformer on the occluded re-ID benchmark, Occluded-DukeMTMC \cite{PGFA}, which contains 15618 training images, 17661 gallery images, and 2210 query images. It is by far the largest and the only occluded person re-ID dataset that contains training set. %It is a new split of DukeMTMC-reID \cite{DukeMTMC-reID} so that its training/query/gallery set contains $9\%/100\%/10\%$ occluded images, respectively.

%{\bf Evaluation metrics.} Following common practices, we use the cumulative matching characteristics (CMC) at Rank-1, Rank-5, and the mean average precision (mAP) to evaluate the performance. All our results are achieved with the single-query mode without re-ranking.

\subsection{Implementation Details and Datasets}

\textbf{The Transformer architecture.}
We use the smallest Vision Transformer model proposed in \cite{ViT}, ViT-Base, as the main architecture of AAformer. It contains 12 Transformer layers with the hidden size of 768 dimensions ($D\!=\!768$). The MLP size is 4 times the hidden size and the head number is 12 for multi-head operations.

\textbf{Data preprocessing.}
The input images are resized to $256\!\times\!128$ and the patch size is $16\!\times\!16$. We adopt the commonly used random cropping \cite{random_earse2}, horizontal flipping and random erasing \cite{Mancs} (with a probability of 0.5) for data augmentation.

\textbf{Training.} 
We warm up the model for 10 epochs with a linearly growing learning rate from $8\!\times\!10^{-4}$ to $8\!\times\!10^{-3}$. Then, the learning rate is decreased with the cosine learning rate decay. It takes 120 epochs to finetune on the re-ID datasets. {AAformer randomly samples 16 identities and 4 images per person to constitute a training batch. The batch size equals to 16*4=64.} SGD optimizer is adopted with a momentum of 0.9 and the weight decay of $1\!\times\!10^{-4}$ to optimize the model. Our methods are implemented on PyTorch and MindSpore\footnote{The codes based on MindSpore will be released at https://gitee.com/typhoonai/AAformer}. The Transformer backbone is pre-trained on ImageNet \cite{ImageNet}.

\begin{table}[h]
\begin{center}
\begin{tabular}{l|c|c|c|c}
\hline
Dataset & \#ID & \#Train & \#Test & \#Images\\ \hline
DukeMTMC-reID & 1402 & 702 & 702 & 36411 \\
Market-1501 & 1501 & 751 & 750 & 32668\\
CUHK03 & 1467 & 767 & 700 & 14097\\
MSMT17 & 4101 & 1041 & 3060 & 126441 \\
\hline
\end{tabular}
\end{center}
\caption{Re-ID datasets and details.}
\label{table:datasets}
\end{table}

\begin{table*}[t]
    \begin{center}
    \begin{tabular}{c|c|cc|cc|cc|cc|cc}
      \hline
      \multirow{3}*{Methods}&\multirow{3}*{Ref}&\multicolumn{2}{|c|}{DukeMTMC}&\multicolumn{2}{|c|}{\multirow{2}*{Market1501}}&\multicolumn{4}{|c}{CUHK03} & \multicolumn{2}{|c}{\multirow{2}*{MSMT17}} \\ \cline{7-10}
      ~&~&\multicolumn{2}{|c|}{-reID}&~&~&\multicolumn{2}{|c}{Labeled}&\multicolumn{2}{|c|}{Detected}&~&~\\ \cline{3-12}
      ~&~&Rank-1&mAP&Rank-1&mAP&Rank-1&mAP&Rank-1&mAP&Rank-1&mAP\\ \hline
      AlignedReID \cite{AlignedReid}& Arxiv18 &-&-&91.8&79.3&-&-&-&-&-&- \\
      PCB+RPP \cite{PCB}& ECCV18&83.3&69.2&93.8&81.6&-&-&63.7&57.5&68.2&40.4\\
      MGN \cite{MGN}&MM18&{88.7}&78.4&\textbf{95.7}&{86.9}&68.0&67.4&66.8&66.0&-&-\\ \hline
      %Pyramid \cite{Pyramid}&{89.0}&{79.0}&95.7&88.2&{78.9}&{76.9}&{78.9}&{74.8}&-&-\\
      %Pyramid \cite{Pyramid} &CVPR19&{89.0}&{94.7}&{79.0}&95.7&{98.4}&88.2&{78.9}&{76.9}&\textcolor{red}{78.9}&\textcolor{red}{74.8}\\
      %GCP \cite{GCP}&{89.7}&78.6&95.2&88.9&77.9&75.6&74.4&69.6&-&-\\
      MSCAN \cite{MSCAN}&CVPR17&-&-&80.8&57.5&-&-&-&-&-&-\\
      PAR \cite{PAR}&ICCV17&-&-&81.0&63.4&-&-&-&-&-&- \\ 
      \hline
      %&MGCAM \cite{MGCAM}&CVPR18&-&-&-&83.8&-&74.3&50.1&50.2&46.7&46.9\\
      SPReID \cite{SPReID}&CVPR18&84.4&71.0&92.5&81.3&-&-&-&-&-&-\\
      PABR \cite{PABR}&ECCV18&84.4&69.3&91.7&79.6&-&-&-&-&-&-\\
      %~&PN-GAN \cite{PN-GAN}&ECCV18&73.6&-&53.2&89.4&-&72.6&-&-&-&-\\
      AANet \cite{AANet}&CVPR19&87.7&74.3&93.9&83.4&-&-&-&-&-&-\\
      %DSA-reID \cite{DSA-reID}&CVPR19&86.2&-&74.3&95.7&-&87.6&{78.9}&{75.2}&{78.2}&{73.1}\\
      $P^2$-Net \cite{P2-Net}&ICCV19&86.5&73.1&95.2&85.6&78.3&73.6&74.9&68.9&-&-\\
      PGFA \cite{PGFA}&ICCV19&82.6&65.5&91.2&76.8&-&-&-&-&-&-\\
      CDPM \cite{CDPM}&TIP19&88.2&77.5&95.2&86.0&75.8&71.1&71.9&67.0&-&-\\
      GASM \cite{GASM}&ECCV20&88.3&74.4&95.3&84.7&-&-&-&-&\underline{79.5}&52.5\\
      PGFA\_{v2} \cite{IPE}&TNNLS21&86.2&72.6&92.7&81.3&-&-&-&-&-&-\\
      \hline
      %DuATM \cite{DuATM}&CVPR18&81.8&90.2&64.6&91.4&97.1&76.6&-&-&-&-\\
      %HA-CNN \cite{HA-CNN}&CVPR18&80.5&-&63.8&91.2&-&75.7&44.4&41.0&41.7&38.6\\
      %Mancs \cite{Mancs}&ECCV18&84.9&-&71.8&93.1&-&82.3&69.0&63.9&65.5&60.5\\
      Non-Local \cite{Non-local}&CVPR18&88.6&{78.7}&94.9&86.8&66.4&65.0&65.3&63.1&76.2&53.3\\
      IANet \cite{IANet}&CVPR19&87.1&73.4&94.4&83.1&-&-&-&-&75.5&46.8\\
      CASN+PCB \cite{CASN}&CVPR19&87.7&73.7&94.4&82.8&73.7&68.0&71.5&64.4&-&-\\
      %~&CAMA \cite{CAMA}&85.8&72.9&94.7&84.5&70.1&66.5&66.6&64.2&-&-\\
      BAT-net \cite{BAT-net}&ICCV19&87.7&77.3&95.1&84.7&{78.6}&{76.1}&\underline{76.2}&\underline{73.2}&{79.5}&{56.8}\\
      JDGL \cite{JDGL}&CVPR19&86.6&74.8&94.8&86.0&-&-&-&-&77.2&52.3\\
      OSNet \cite{OSNet}&ICCV19&88.6&73.5&94.8&84.9&-&-&72.3&67.8&78.7&52.9\\
      CI-CNN \cite{CI-CNN}&TIP19&87.6&81.3&94.2&83.9&-&-&-&-&-&-\\
      OCLSM \cite{center}&TIP20&87.7&79.0&94.6&87.4&-&-&-&-&78.8&57.0\\
      ISP \cite{ISP}&ECCV20&89.6&80.0&95.3&\underline{88.6}&76.5&74.1&75.2&71.4&-&-\\
      PAT \cite{PAT}&CVPR21&{88.8}&78.2&\underline{95.4}&{88.0}&-&-&-&-&-&-\\
      PFE \cite{PFE}&TIP21&88.2&75.9&95.1&86.2&-&-&-&-&79.1&52.3\\
      FA-Net \cite{FA-Net}&TIP21&88.7&77.0&95.0&84.6&-&-&-&-&76.8&51.0\\
      HOReID \cite{HOReID}&TIP21&88.1&{79.8}&\textbf{95.7}&\textbf{88.7}&-&-&-&-&74.4&50.4\\
      TransReID$^{-}$ \cite{transreid}& ICCV21&\underline{89.6}&{79.8}&94.2&87.7&\underline{78.9}&\underline{76.9}&75.1&72.9&{82.5}&{63.6}\\
      reID-NAS \cite{reID-NAS}&TNNLS21&88.1&74.6&95.1&85.7&-&-&-&-&79.5&53.3\\
      MHSA-Net \cite{MHSA-Net}&TNNLS22&87.3&73.1&94.6&84.0&75.6&72.7&72.8&69.3&-&-\\
      
      DCAL \cite{DCAL}&CVPR22&89.0&\underline{80.1}&94.7&87.5&-&-&-&-&\underline{83.1}&\underline{64.0}\\
      %NFormer \cite{NFormer}&&&&&&&&&&\\
      %RGA \cite{RGA}&-&-&-&-&-&-&-&-&-&-\\
      %SCSN \cite{SCSN}&-&-&-&-&-&-&-&-&-&-\\
      %MHN-6 \cite{MHN}&ICCV19&89.1&94.6&77.2&95.1&98.1&85.0&77.2&72.4&71.7&65.4\\
      %~& \cite{ABD-Net}&ICCV19&{89.0}&-&78.6&\textcolor{red}{96.6}&-&{88.8}&-&-&-&-\\
      %SCAL \cite{SCAL} &ICCV19 &89.0&95.1&79.6&{95.8}&98.5&{88.9}&74.8&72.3&71.1&68.6 \\
      \hline
      %~&baseline&~&87.7&94.3&77.2&94.3&97.9&87.0&~&~&~&~\\
      %AAformer (ours)&~&\textcolor{red}{89.7}&\textcolor{red}{-}&\textcolor{red}{79.7}&\textcolor{red}{96.0}&\textcolor{red}{98.8}&\textcolor{red}{89.4}&\textcolor{red}{80.3}&\textcolor{red}{77.2}&{77.1}&{74.3} \\
      ViT-baseline & - & 88.3 & 78.5 & 94.2 & 86.3 & 75.3 & 74.9 & 74.0 & 71.6 & 79.7 & 58.9 \\
      AAformer (ours)&this paper&\textbf{90.1}&\textbf{80.9}&\underline{95.4}&{88.0}&\textbf{80.3}&\textbf{79.0}&\textbf{78.1}&\textbf{77.2}&\textbf{84.4}&\textbf{65.6} \\
      \hline
    \end{tabular}
    \end{center}
    \caption{Comparison with the state-of-the-art methods for the holistic person re-ID problem. Methods in the 1st group are the stripe-based methods. Methods in the 2nd group are bounding box-based methods. Methods in the 3rd group are extra semantics-based methods. Methods in the 4th group include the general attention methods of CNN \cite{IANet, CASN}, the work that combines self-attention with CNN \cite{Non-local, PAT} and the method based on pure Transformer \cite{transreid,DCAL}. TransReID$^{-}$ means the side information is removed for a fair comparison.
    % The best results in each category are shown in bold. \multirow{2}*{DukeMTMC-reID}
    }
    \label{table:state_of_the_art}
\end{table*}

\textbf{Datasets and Evaluation Metrics.} We conduct experiments on four widely used person re-ID benchmarks, i.e., DukeMTMC-reID\cite{DukeMTMC-reID}, Market-1501 \cite{Market1501}, CUHK03 (New Protocol) \cite{CUHK03-2}, and the large-scale MSMT17 \cite{MSMT17}. The standard training/test ID split is used and detailed in Table~\ref{table:datasets}.  Following common practices, we use the cumulative matching characteristics (CMC) and the mean average precision (mAP) to evaluate the performance. Euclidean distance is used to measure the feature distances.

\subsection{Comparison with State-of-the-art Methods}

\begin{table}
\begin{center}
\begin{tabular}{l|cccc}
\hline%\noalign{\smallskip}
Methods & Rank-1 & Rank-5  & mAP\\
%\noalign{\smallskip}
\hline
%HACNN \cite{HACNN} & 34.4 & 51.9  & 26.0 \\
%Adver Occluded \cite{adver_occluded} & 44.5 & - & 32.2 \\
PCB \cite{PCB} & 42.6 & 57.1 & 33.7 \\
Part Bilinear \cite{part_bilinear} & 36.9 & -  & - \\
FD-GAN \cite{FD-GAN} & 40.8 & -  & - \\
DSR \cite{DSR} & 40.8 & 58.2  & 30.4 \\
SFR \cite{SFR} & 42.3 & 60.3  & 32.0 \\
PGFA \cite{PGFA} & 51.4 & 68.6  & 37.3 \\
PGFA\_v2 \cite{IPE} & 56.3 & 72.4 & 43.5 \\
HOReID \cite{HOReID} & 55.1 & - & 43.8 \\
ISP \cite{ISP} & 62.8 & 78.1 & 52.3\\
PAT \cite{PAT} & 64.5 & - & 53.6 \\
\hline
ViT-baseline & 60.8 & 77.8 & 52.5 \\
AAformer (ours) & \textbf{67.1} & \textbf{81.6} & \textbf{58.2} \\
\hline
\end{tabular}
\end{center}
\caption{Comparison with the state-of-the-art methods for the occluded re-ID problem on Occluded-DukeMTMC.}
\label{table:state_of_the_art_occluded}
\end{table}

We compare our method with the state-of-the-art methods on four widely used holistic benchmarks and one occluded benchmark in Table~\ref{table:state_of_the_art} and Table~\ref{table:state_of_the_art_occluded}. We also show the results of the baseline model (ViT-baseline \cite{ViT}) in the tables. {We follow BoT \cite{baseline}, a popular baseline method in person ReID filed, to form the ViT-baseline. Most of the settings refer to BoT, including the warmup learning rate, random erasing augmentation [34], label smoothing [48] and BNNeck [46].} %The baseline is ViT-base model \cite{ViT} trained with re-ID settings and We also show the results of ViT-baseline in the tables.

\textbf{DukeMTMC-reID.} AAformer obtains the best results and outperforms others by considerable margins, e.g., at least $0.5\%$ on Rank-1 accuracy and $0.8\%$ on mAP accuracy. The compared methods are categorized into four groups, i.e., stripe-based methods, bounding box-based methods, extra semantics-based methods, and others. The others include the general attention methods based on CNN \cite{IANet, CASN}, the work that combines self-attention with CNN \cite{Non-local, PAT} and the method based on generative adversarial network \cite{JDGL}. AAformer surpasses all the above methods by considerable margins, which validates the superiority of locating both human parts and non-human ones.

\textbf{Market1501.} AAformer achieves state-of-the-art performance. Specifically, AAformer obtains the second-best results and is only slightly behind the firsts \cite{HOReID}. As the performance on this dataset is nearly saturated, the results of sate-of-the-art methods are very close.

\textbf{CUHK03.} AAformer obtains the best performance on both labeled and detected sets. More specifically, AAformer outperforms the second-best algorithms \cite{transreid,BAT-net} by $1.4\%/2.1\%$ and $1.9\%/4.0\%$ on labeled and detected sets with regards to Rank-1 and mAP, respectively. Compared with ISP \cite{ISP}, which is a CNN-based method that also conducts adaptive part alignment, AAformer obtains much better performance on this difficult dataset. We owe this to our online clustering, which can be more accurate than the off-line clustering in ISP. AAformer also significantly surpasses the ViT-baseline by $4.1\!\sim\!5.6\%$, which validates the effectiveness of MAA.
 
\textbf{MSMT17.} On the largest dataset, AAformer also outperforms the state-of-the art results. On the metric of Rank-1, AAformer outperforms the second-best \cite{transreid} by $1.3\%$. On the metric of mAP, AAformer outperforms the second-best \cite{transreid} by $1.6\%$. It should be noted that we remove the side information in \cite{transreid} for fair comparison. Besides, we strongly recommend not using the side information like camera IDs. Because if the camera IDs are used, the trained model can not be applied to other monitoring systems as the cameras are different, which largely limits the generality and practicability.

\textbf{Occluded-DukeMTMC.} AAformer sets the new state-of-the-art performance and outperforms others by a large margin, at least $2.6\%$/$4.6\%$ in Rank-1/mAP. Owing to the Auto-Alignment, the {[PART]}s can adaptively focus on the visible areas in the occluded images and extract discriminative features from these visible areas. We also visualize the attention maps of {[PART]}s in the ablation studies to validate this.

\subsection{Ablation Studies}

\textbf{The effectiveness of the {[PART]}s.}
We first conduct experiments to validate the effectiveness of {[PART]}s. We begin by naively adopting the alignment paradigm of CNNs, i.e., part masks with part pooling, to the baseline Transformer model (ViT-Base). Three typical methods, i.e., PCB, MGN, and SPReID, are used for this experiment. The first two methods are stripe-based methods and the third method adopts the extra semantics.
As shown in Table~\ref{table:ablation}, naively applying the alignment paradigm of CNNs even reduces the performance of the baseline model, which may be because pooling is not suitable for Transformer-based person re-ID. %As analyzed before, we believe it is the pooling operation that does not match the Transformer.

Next, we discard the pooling operation and add the proposed {[PART]}s to ViT to extract part features.
These typical CNN-based methods are easily transferred to Transformer by our {[PART]}s, 
{e.g., we assign each {[PART]} with patches within a stripe to simulate the image partition of PCB and divide the image patches according to a pre-trained human parsing model to simulate SPReID.} {Specifically, in the simulation experiment of SPReID, if a patch contains multiple semantics, we assign the patch to all the semantics it contains.}
As shown in Table~\ref{table:ablation}, models with {[PART]}s bring consistent improvement to the baseline, e.g., {[PART]}s with MGN's partitioning surpasses baseline by $2.8\%/4.5\%$ in terms of Rank-1/mAP on MSMT17, validating the significant advantages of {[PART]}s in extracting part features for Transformer. {This is because [PART]s can extract the local features through the self-attention mechanism in every layer of Transformer, while the part pooling operation only obtains the local features through the last layer.}

\begin{table}[t]
\begin{center}
\begin{tabular}{l|c|cc|cc}
\hline%\noalign{\smallskip}
\multirow{2}*{Methods}&\multirow{2}*{Scheme}&\multicolumn{2}{c|}{MSMT17}&\multicolumn{2}{c}{{CUHK03(L)}}\\  \cline{3-6}
 &  & Rank-1 & mAP & Rank-1 & mAP\\
%\noalign{\smallskip}
\hline
ViT-baseline &None& 79.7  & 58.9  & 75.3 & 74.9 \\
\hline
+PCB & Pool & 81.3  & 59.8  & 76.3 & 75.6 \\
+MGN & Pool & 81.6 & 60.5  & 76.5 & 75.3 \\
+SPReID & Pool & 79.6  & 58.7  & 74.9 & 73.7 \\
\hline
+PCB & PT & 82.2  & 62.1  & 78.4 & 76.9 \\
+MGN & PT & 82.5  & 63.4  & 78.8 & 77.3 \\
+SPReID & PT & 81.2  & 60.3 & 79.1 & 77.5 \\
\hline
ViT-baseline & 6*CLS & 80.2  & 59.4  & 75.7 & 75.1 \\
\hline
%AAformer& PT & \textbf{83.6} & \textbf{63.2} & \textbf{79.9} & \textbf{77.8} \\
AAformer& PT & \textbf{84.4} & \textbf{65.6} & \textbf{80.3} & \textbf{79.0} \\
\hline
\end{tabular}
\end{center}
\caption{The effectiveness of {[PART]}s and AAformer. ``Pool'' means ``Pooling'' and ``PT'' means ``{[PART]}s''.}
\label{table:ablation}
\end{table}

Besides, to prove the improvement is not brought by the increase of feature dimension, we conduct the experiment which directly adds more CLS tokens to ViT. {The results in Table~\ref{table:ablation} validate that naively increasing the feature dimension only slightly improves the performance} as the information contained in these CLS tokens is similar and redundant.

\begin{figure}[t]
\begin{center}
\includegraphics[width=1.\linewidth]{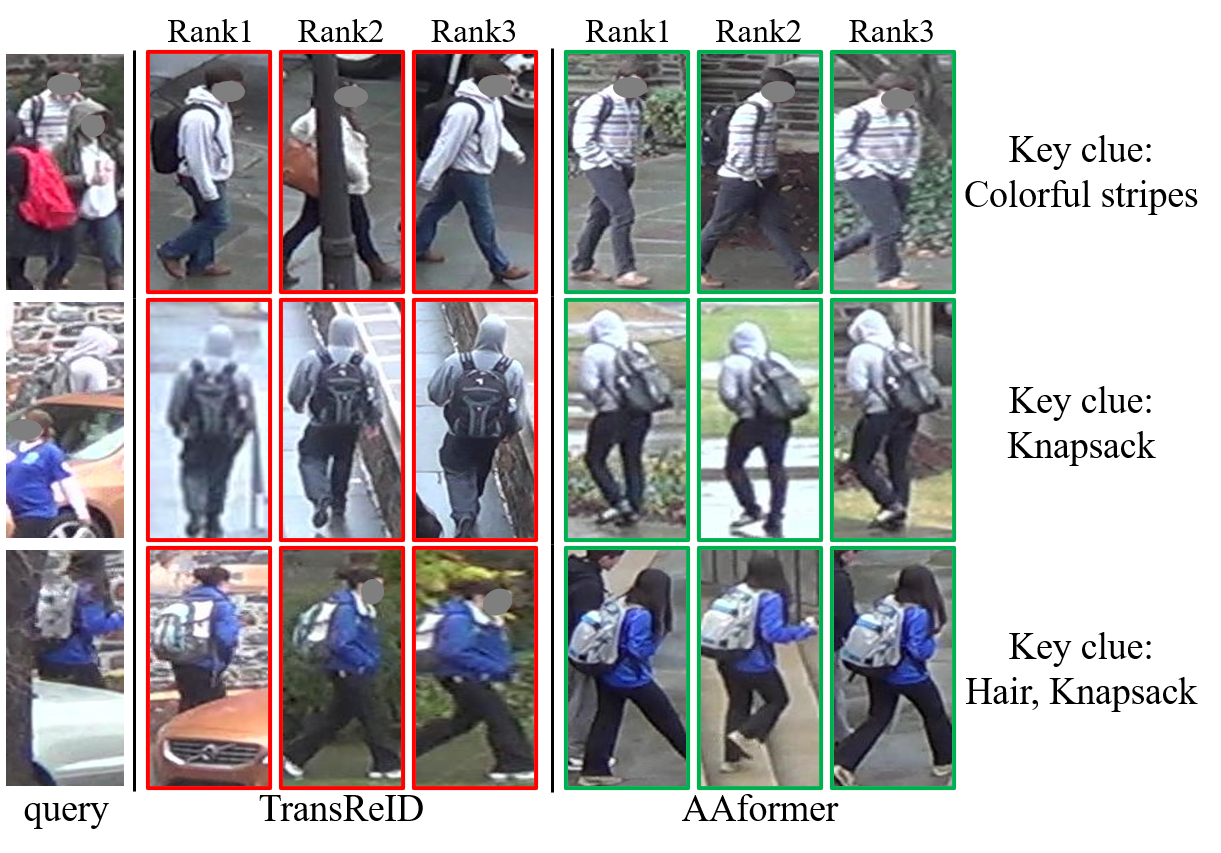}
\end{center}
   \caption{The ranking lists of TransReID and AAformer in misalignment scenes. Tiny clues are found by our AAformer.}
\label{fig:ranking_list}
\end{figure}

\textbf{The effectiveness of AAformer.}   
As shown in the last row of Table~\ref{table:ablation}, AAformer surpasses all the typical alignment methods reimplemented on {[PART]}s by large margins, e.g., AAformer outperforms {[PART]}s with MGN's partitioning by $1.9\%/2.2\%$ in terms of Rank-1/mAP on MSMT17, which validates the superiority of online adaptive patch assignment over the fixed patterns. Besides, to validate the strong ability of AAformer in finding the tiny clues in the misalignment scenes, we compare the ranking list between TransReID \cite{transreid} and AAformer in Figure~\ref{fig:ranking_list}. As shown, AAformer can find the discriminative tiny clues from both human parts and non-human ones, {while TransReID fails at this due to the lacking of an alignment scheme.}

\textbf{Optimal Transport \emph{vs.} Nearest Neighbor.} Most other clustering methods (K-means, DBSCAN) cannot be used for adaptive patch assignment, because they cannot guarantee the grouped clusters of different images (or the clusters of the same image from different layers) to have the same semantics and correspond to the consistent {[PART]}s. A naive alternative algorithm is the Nearest Neighbor (NN), which assigns the patch embedding to the nearest {[PART]}. Unfortunately, in our experiments, we observe that most patch embeddings are grouped to the same {[PART]} when adopting NN as there is no constraint on the patch assignment. The results in Table~\ref{table:OT_NN} also demonstrate the superiority of Optimal Transport over Nearest Neighbor.
%As we use Optimal Transport \cite{OT} for patch assignment, we also test the Nearest Neighbor (NN) algorithm \cite{NN} for the automatic patch assignment. In the setting of AAformer with NN, we compute the distance between image patches and part tokens and every image patch is assigned to its nearest part token as the part assignment. As shown in Table~\ref{}, the NN algorithm

\begin{table}[h]
\begin{center}
\begin{tabular}{c|cc|cc}
\hline
\multirow{2}*{Settings}&\multicolumn{2}{|c|}{MSMT17}&\multicolumn{2}{|c}{CUHK03(L)}\\ \cline{2-5}
~ & Rank-1 & mAP & Rank-1 & mAP\\
\hline
Nearest Neighbor & 82.2 & 61.5 & 79.1 & 77.4 \\
Optimal Transport  & \textbf{84.4} & \textbf{65.6} & \textbf{80.3} & \textbf{79.0} \\
\hline
\end{tabular}
\end{center}
\caption{The comparison of patch assignment with the Nearest Neighbor and the Optimal Transport algorithm.}
\label{table:OT_NN}
\end{table}

{\textbf{Different numbers of [PART]s in AAformer.} Intuitively, the number of [PART]s $P$ determines the granularity of the body parts. Thanks to the flexibility of AAformer, we can partition the input image with different granularity in parallel in an AAformer model. For example, granularity $\{2, 3\}$ indicates there are 5 [PART]s added. The first two [PART]s divide the image patches into two groups and the last three [PART]s divide them into three. All the [PART]s are randomly and independently initialized, but they will be assigned image patches of different semantics in the training, and eventually learn to be different part prototypes. The results of AAformer with different numbers of [PART]s are shown in Table~\ref{table:gran}. As we can observe, the setting of $\{2,3\}$ usually obtains the best accuracy, which is consistent with the setting of MGN. Besides, the results also show that the number of [PART]s is not the more the better. All the previous results are obtained with the default setting of granularity $\{2,3\}$.}

\begin{table}[h]
\begin{center}
\begin{tabular}{c|cc|cc}
\hline
\multirow{2}*{$P$}&\multicolumn{2}{|c|}{MSMT17}&\multicolumn{2}{|c}{CUHK03(L)}\\ \cline{2-5}
~ & Rank-1 & mAP & Rank-1 & mAP\\
\hline
\{4\} & 82.4  & 62.3  & 78.5 & 76.9 \\
\{5\} & 82.0  & 61.6  & 79.8 & 78.2 \\
\{6\} & 81.7  & 61.4 & 79.2 & 77.3 \\
%\{2,3\}  &\textbf{83.6} & \textbf{63.2}& \textbf{79.9} & \textbf{77.8} \\
\{2,3\}  &\textbf{84.4} & \textbf{65.6}& \textbf{80.3} & \textbf{79.0} \\
\{3,4\} & 83.1  & 63.2 & 79.4 & 78.0 \\
\{2,3,4\} & 82.9 & 63.0  & 78.8 & 77.6 \\
\hline
\end{tabular}
\end{center}
\caption{{Ablation studies on different numbers of [PART]s}.}
\label{table:gran}
\end{table}

\begin{figure*}[h!]
\begin{center}
\includegraphics[width=0.95\linewidth]{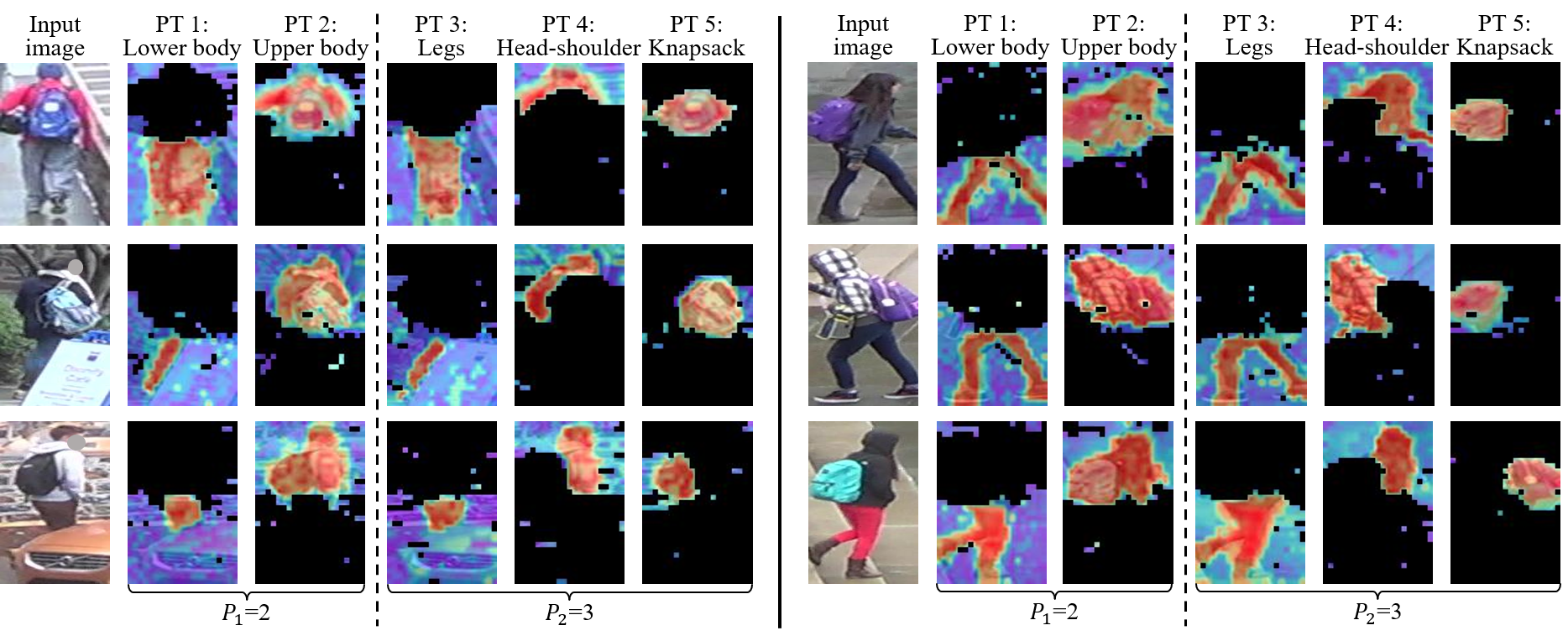}
\end{center}
   \caption{{The attention map of {[PART]} (PT). For one [PART], the patches not assigned to it are masked by black. The color range from blue to red indicates increasing attention.}}
\label{fig:visualization}
\end{figure*}

\textbf{Visualization of AAformer.} Lastly, we conduct the visualization experiments to show the focusing areas of {[PART]}s. {We visualize the attention map of {[PART]}s in the first head of the third Transformer layer,} which is shown in Figure~\ref{fig:visualization}. As we can observe, AAformer can focus on both human parts and non-human ones, which are both crucial for re-ID. This is also the main superiority of AAformer over existing methods. {Besides, we can also find that AAformer can effectively handle the pose variation and the occlusion problems.}
Owing to the self-attention mechanism, the {[PART]}s have very low responses to the background patches, thus it is no need for AAformer to remove the background patches. The visualization also proves the semantics consistency of the located parts among images.% which is analysed in the end of Sect. 3.2. %Figure~\ref{fig:visualization} also shows the part tokens have very low responses to the background patches owing to the self-attention mechanism. % thus it is no need to remove these noise patches in advance. Actually, these noise patches can be easily removed by setting a similarity threshold in the clustering, i.e., if the maximum similarity between a patch and part tokens is less than a threshold, this patch is considered as the background. 
%It should be noted that we do not need to guarantee the semantic consistency of images of different persons. As long as the semantics of images of the same person is consistent, AAformer can retrival the correct images from the gallery.

\section{Conclusion}
In this paper, we propose the {[PART]}s  for Transformer to extract the part-level features. We propose to integrate the part localization process into the self-attention of Transformer to online learn the part representations. The proposed method, Auto-Aligned Transformer (AAformer), can adaptively locate both human parts and non-human ones using Optimal Transport (OT) without the help of extra semantics. It should be noted that most other clustering methods (k-means, DBSCAN) cannot be used in AAformer, because they cannot guarantee that the grouped clusters of different images (or the same image from different layers) are of the identical semantics and correspond to the consistent {[PART]}s. It is our novel design of AAformer that makes it possible to use OT to solve the adaptive alignment problem. Extensive experiments also validate the effectiveness of {[PART]}s and the superiority of AAformer over lots of state-of-the-art methods.

% or


%\appendices

%\section{}

%\section*{Acknowledgment}
%This work was supported by Key-Area Research and Development Program of Guangdong Province No.2021B0101410003, and National Natural Science Foundation
%of China under Grants 62276260, 62002356, 61976210.


\ifCLASSOPTIONcaptionsoff
  \newpage
\fi



%\bibliographystyle{IEEEtran}
\bibliographystyle{IEEEtran}
\bibliography{bibtex/IEEEabrv,reference}

\begin{IEEEbiography}[{\includegraphics[width=1in,height=1.25in,clip,keepaspectratio]{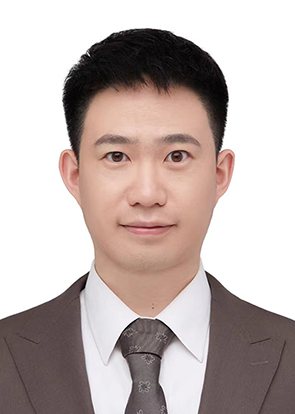}}]{Kuan Zhu}
received the B.E. degree from Xiamen University, Xiamen, China, in 2018, and the Ph.D. degree in pattern recognition and intelligence systems from the National Laboratory of Pattern Recognition, Institute of Automation, Chinese Academy of Sciences, in 2023. He is currently an Assistant Professor with the Institute of Automation, Chinese Academy of Sciences. His current research interests include self-supervised learning, open-world visual recognition, large language model and multimodal model.

Multimodal model.
\end{IEEEbiography}

\begin{IEEEbiography}[{\includegraphics[width=1in,height=1.25in,clip,keepaspectratio]{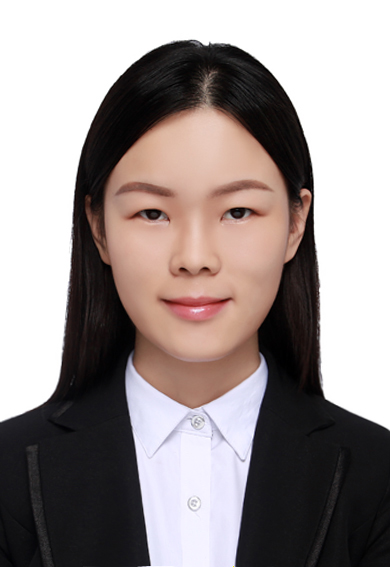}}]{Haiyun Guo}
received the B.E. degree from Wuhan University in 2013 and the Ph.D. degree in pattern recognition and intelligence systems from the Institute of Automation, University of Chinese Academy of Sciences, in 2018. She is currently an Associate Researcher with the Institute of Automation, Chinese Academy of sciences. Her current research interests
include pattern recognition and machine learning, image and video processing, and intelligent video surveillance.
\end{IEEEbiography}

\begin{IEEEbiography}[{\includegraphics[width=1in,height=1.25in,clip,keepaspectratio]{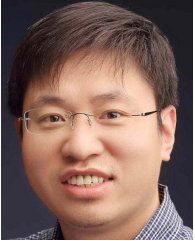}}]{Shiliang Zhang}  received the Ph.D. degree in computer science from the Institute of Computing Technology, Chinese Academy of Sciences. He was a Post-Doctoral Scientist with NEC Laboratories America and a Post-Doctoral Research Fellow with The University of Texas at San Antonio. He is currently an Associate Professor with Tenure with the Department of Computer Science, School of Electronic Engineering and Computer Science, Peking University.

His research interests include large-scale image retrieval and computer vision. He has authored or co-authored over 100 papers in journals and conferences, including IJCV, IEEE Trans. on PAMI, IEEE Trans. on Image Processing, IEEE Trans. on NNLS, IEEE Trans. on Multimedia, ACM Multimedia, ICCV, CVPR, ECCV, NeurIPS, AAAI, IJCAI, etc. He was a recipient of the Outstanding Doctoral Dissertation Awards from the Chinese Academy of Sciences and Chinese Computer Federation, the President Scholarship from the Chinese Academy of Sciences, the NEC Laboratories America Spot Recognition Award, the NVidia Pioneering Research Award, and the Microsoft Research Fellowship. He served as the Associate Editor (AE) of Computer Vision and Image Understanding (CVIU) and IET Computer Vision, Guest Editor of ACM TOMM, and Area Chair of CVPR, AAAI, ICPR, and VCIP. His research is supported by the The National Key Research and Development Program of China, Natural Science Foundation of China, Beijing Natural Science Foundation, and Microsoft Research, etc.

\end{IEEEbiography}

\begin{IEEEbiography}[{\includegraphics[width=1in,height=1.25in,clip,keepaspectratio]{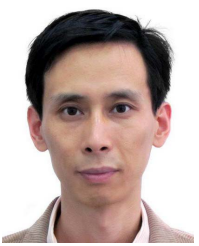}}]{Yaowei Wang} received his Ph.D. degree in computer science from the University of Chinese Academy of Sciences in 2005. He is currently a professor with the Peng Cheng Laboratory, Shenzhen, China. He is the author or co-author of more than 120 technical articles in international journals and conferences, including TOMM, ACM MM, IEEE TIP, CVPR, ICCV, and IJCAI. His current research interests include multimedia content analysis and understanding, machine learning, and computer vision. He serves as the chair of the IEEE Digital Retina Systems Working Group and a member of IEEE, CIE, CCF, CSIG. He was the recipient of the second prize of the National Technology Invention in 2017, the first prize of the CIE Technology Invention in 2015, and the first prize of the CIE Scientific and Technological Progress in 2022.

\end{IEEEbiography}

\begin{IEEEbiography}[{\includegraphics[width=1in,height=1.25in,clip,keepaspectratio]{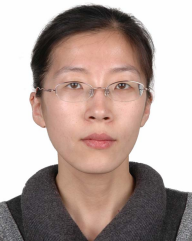}}]{Jing Liu} received the B.E. and M.S. degrees from Shandong University, Shandong, in 2001 and 2004, respectively, and the Ph.D. degree from the Institute of Automation, Chinese Academy of Sciences, Beijing, in 2008. She is currently a Professor with the Institute of Automation, Chinese Academy of Sciences. Her current research interests include deep learning, image content analysis and classification, and multimedia understanding and retrieval.
\end{IEEEbiography}

\begin{IEEEbiography}[{\includegraphics[width=1in,height=1.25in,clip,keepaspectratio]{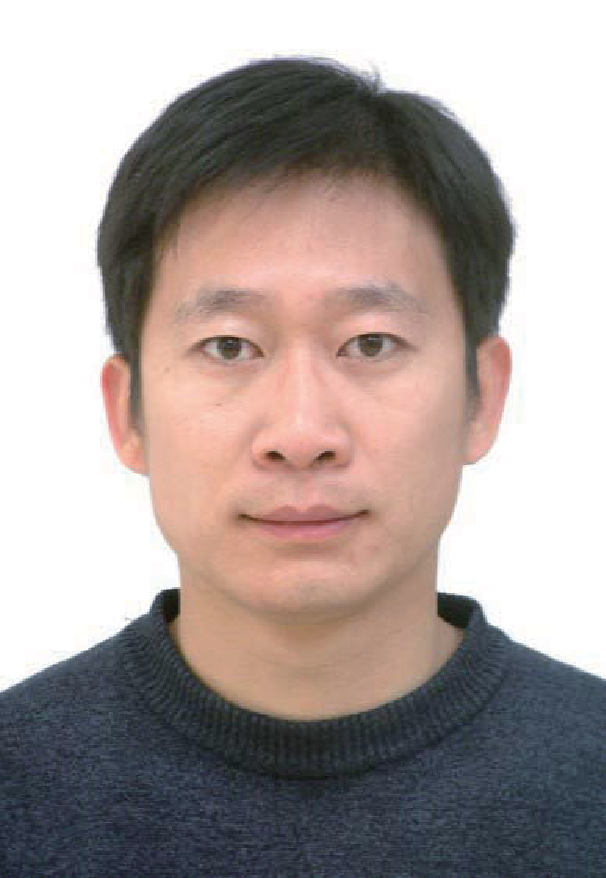}}]{Jinqiao Wang}
received the B.E. degree from the Hebei University of Technology, China, in 2001, the M.S. degree from Tianjin University, China, in 2004, and the Ph.D. degree in pattern recognition and intelligence systems from the National Laboratory of Pattern Recognition, Chinese Academy of Sciences, in 2008. He is currently a Professor with the Institute of Automation, Chinese Academy of Sciences. His research interests include pattern recognition and machine learning, large multimodal model, Image and video analysis, object detection and recognition.
\end{IEEEbiography}

\begin{IEEEbiography}[{\includegraphics[width=1in,height=1.25in,clip,keepaspectratio]{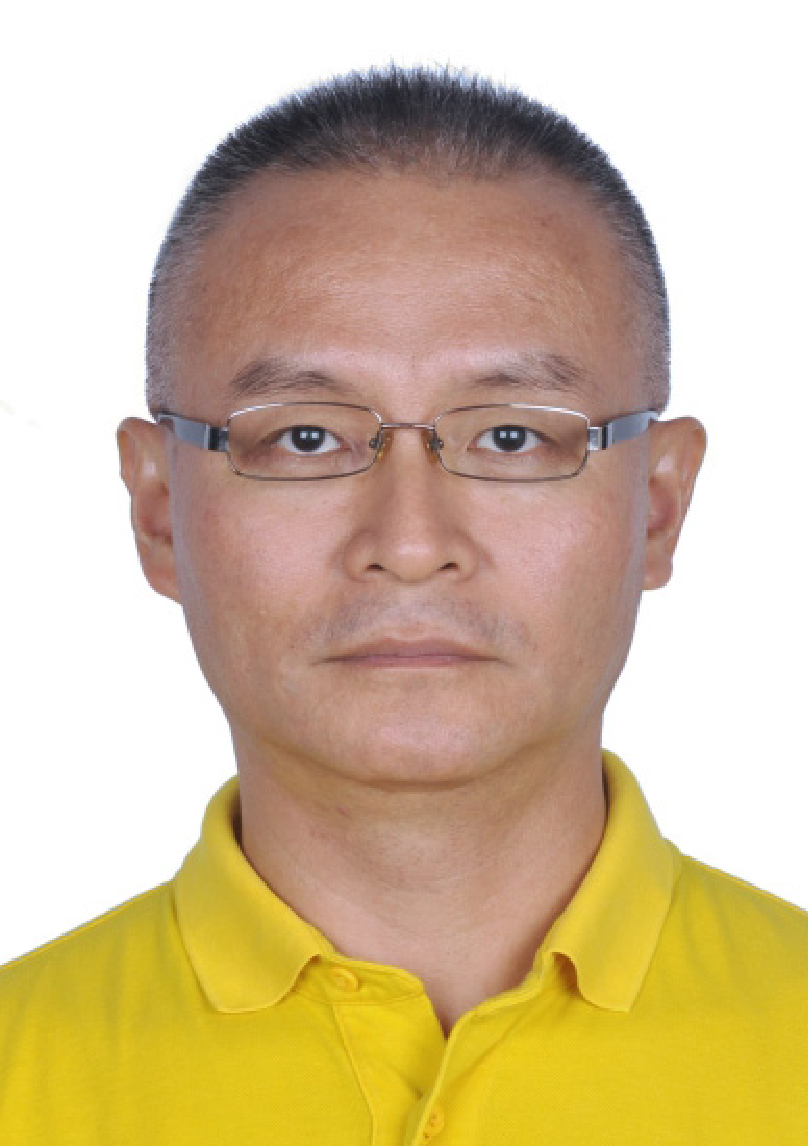}}]{Ming Tang}
received the B.S. degree in computer science and engineering and M.S. degree in artificial intelligence from Zhejiang University, Hangzhou, China, in 1984 and 1987, respectively, and the Ph.D. degree in pattern recognition and intelligent system from the Chinese Academy of Sciences, Beijing, China, in 2002. He is currently a Professor with the Institute of Automation, Chinese Academy of Sciences. His current research interests include computer vision and machine learning.
\end{IEEEbiography}


%\end{IEEEbiography}


%\newpage



%\vfill


\end{document}